\def\aaaianonymous{false}
\documentclass[letterpaper]{article} 

\usepackage{aaai2026} 

\usepackage{times}  
\usepackage{helvet}  
\usepackage{courier}  
\usepackage[hyphens]{url}  
\usepackage{graphicx} 
\usepackage{natbib}  
\usepackage{caption} 
\usepackage{algorithm}
\usepackage{algorithmic}
\usepackage{enumitem}

\usepackage{newfloat}
\usepackage{listings}
\usepackage{amsmath}
\usepackage{multirow}
\usepackage{booktabs} 
\usepackage{makecell}
\usepackage{amsfonts}
\DeclareCaptionStyle{ruled}{labelfont=normalfont,labelsep=colon,strut=off} 
\floatstyle{ruled}
\newfloat{listing}{tb}{lst}{}
\floatname{listing}{Listing}

\ifdefined\aaaianonymous
    \title{Any-Optical-Model: A Universal Foundation Model for Optical Remote Sensing}
\else
    \title{AAAI Press Formatting Instructions \\for Authors Using \LaTeX{} --- A Guide}
\fi

\author{
    Xuyang Li\textsuperscript{\rm 2,3},
    Chenyu Li\textsuperscript{\rm 1}\thanks{Corresponding author.},
    Danfeng Hong\textsuperscript{\rm 1}
}
\affiliations{
    \textsuperscript{\rm 1}Southeast University, 211189 Nanjing, China \\
    \textsuperscript{\rm 2}Aerospace Information Research Institute, Chinese Academy of Sciences, 100094 Beijing, China\\
    \textsuperscript{\rm 3}School of Electronic, Electrical and Communication Engineering, \\
    University of Chinese Academy of Sciences, 100049 Beijing, China \\

    lixuyang23@mails.ucas.ac.cn, chenyuli.erh@gmail.com, danfeng.hong@seu.edu.cn
}

\usepackage{bibentry}

\begin{document}

\maketitle

\begin{abstract}
Optical satellites, with their diverse band layouts and ground sampling distances, supply indispensable evidence for tasks ranging from ecosystem surveillance to emergency response. However, significant discrepancies in band composition and spatial resolution across different optical sensors present major challenges for existing Remote Sensing Foundation Models (RSFMs). These models are typically pretrained on fixed band configurations and resolutions, making them vulnerable to real world scenarios involving missing bands, cross sensor fusion, and unseen spatial scales, thereby limiting their generalization and practical deployment.
To address these limitations, we propose Any Optical Model (AOM), a universal RSFM explicitly designed to accommodate arbitrary band compositions, sensor types, and resolution scales. To preserve distinctive spectral characteristics even when bands are missing or newly introduced, AOM introduces a spectrum-independent tokenizer that assigns each channel a dedicated band embedding, enabling explicit encoding of spectral identity. To effectively capture texture and contextual patterns from sub-meter to hundred-meter imagery, we design a multi-scale adaptive patch embedding mechanism that dynamically modulates the receptive field. Furthermore, to maintain global semantic consistency across varying resolutions, AOM incorporates a multi-scale semantic alignment mechanism alongside a channel-wise self-supervised masking and reconstruction pretraining strategy that jointly models spectral-spatial relationships.
Extensive experiments on over 10 public datasets, including those from Sentinel-2, Landsat, and HLS, demonstrate that AOM consistently achieves state-of-the-art (SOTA) performance under challenging conditions such as band missing, cross sensor, and cross resolution settings. These results highlight AOM as a crucial step toward building truly general-purpose RSFMs.
\end{abstract}

\ifdefined\aaaianonymous
\else
\begin{links}
    \link{Code}{https://aaai.org/example/code}
    \link{Datasets}{https://aaai.org/example/datasets}
    \link{Extended version}{https://aaai.org/example/extended-version}
\end{links}
\fi

\ifdefined\aaaianonymous

\section{Introduction}
\begin{figure}[ht]
      \centering	   
      \includegraphics[width=0.47\textwidth]{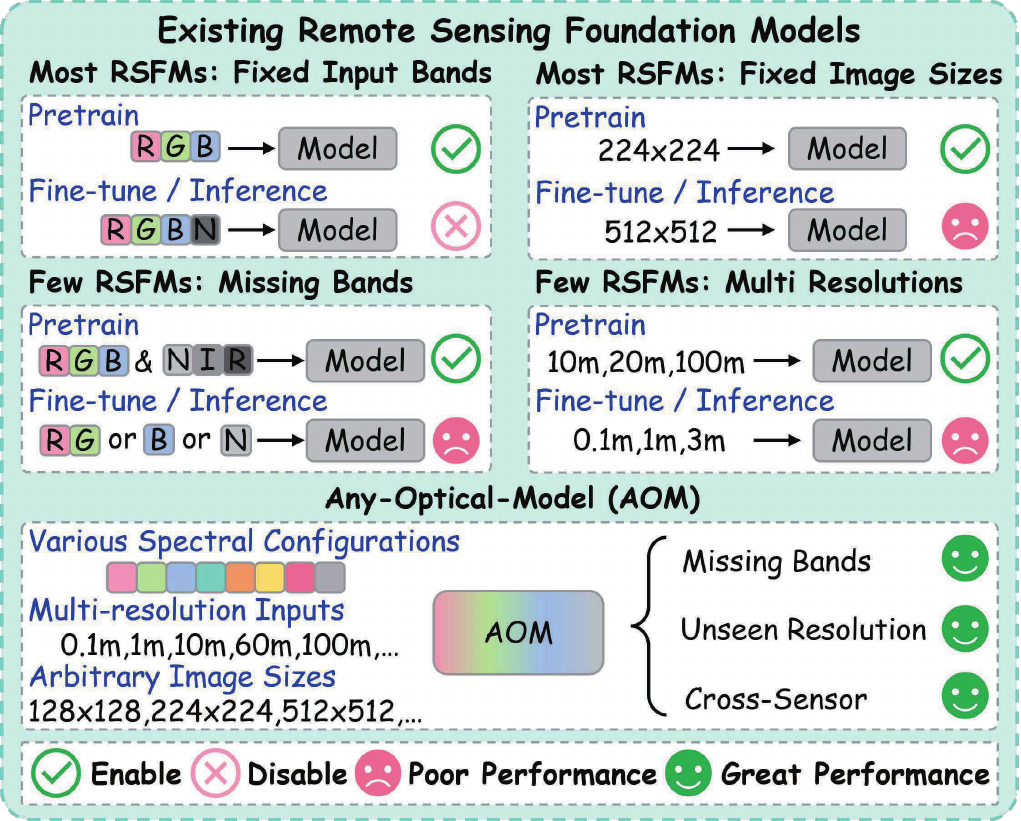}
      \caption{\textbf{Limitations of current remote sensing foundation models.} Mainstream models fail to adapt to varying spectral bands, spatial resolutions, and image sizes across pretraining and downstream tasks.}
\label{fig:motivation}
\end{figure}
Spectral remote sensing (RS) data \cite{hong2025hyperspectral} is a cornerstone of optical Earth observation, leveraging its ability to capture rich spectral and spatial information to deliver significant value in applications such as environmental monitoring, disaster response, and precision agriculture. Compared to single RGB imagery, spectral data offers higher spectral resolution and inherently exhibits two key characteristics: band heterogeneity and scale diversity. Different satellites vary significantly in band count, central wavelength, and bandwidth, while spatial resolutions span from sub-meter to hundred-meter scales, with differences reaching orders of magnitude~\cite{li2024s2mae}. In recent years, the integration of artificial intelligence in the RS field has led to the development of various optical foundation models to support downstream tasks such as land cover classification, change detection, and object recognition. However, existing RSFMs are typically pretrained on fixed band configurations and spatial scales, implicitly assuming complete band availability and constant resolution. This assumption limits their performance in real-world scenarios involving (1) missing or additional bands, (2) cross-satellite data, or (3) unseen resolutions, severely constraining their generalization and practical utility, as described in Figure~\ref{fig:motivation}.

Current mainstream RSFMs~\cite{guo2024skysense,astruc2025anysat} typically process multispectral data by treating it as a monolithic input, overlooking differences in physical properties and complex inter-band interactions. Some models, such as DOFA~\cite{xiong2024neural} and SpectralGPT~\cite{hong2024spectralgpt}, incorporate sophisticated designs that enable their structures to accommodate data inputs from different channels. However, in real-world scenarios, spectral data often exhibits significant heterogeneity due to sensor variations or missing bands. For instance, when certain bands are unavailable, models like DOFA struggle to effectively extract features from the remaining bands, resulting in degraded performance.
Additionally, most existing models are designed for specific resolutions, employing single-scale patch embedding that fails to adapt to the texture and contextual modeling required for unseen resolutions. RS imagery spans a vast range of resolutions, from sub-meter to hundreds of meters. While recent studies have attempted to address scale variability, they primarily focus on high-resolution RGB imagery and do not adequately account for the inter-band spatial feature correlations and sensor-specific differences inherent in multispectral data~\cite{reed2023scale,noman2024rethinking}. Scale variations can also disrupt global semantic understanding, leading to unstable performance in multi-scale scenarios. Consequently, models often require retraining or become ineffective when processing data from different sensors or resolutions. \textbf{How can we design a model that seamlessly adapts to arbitrary band compositions, sensor types, and resolution scales?}

To overcome these limitations, we introduce Any-Optical-Model (\textbf{\emph{AOM}}), an RSFM that unifies spectral and spatial feature learning within a single framework. First, to tolerate missing or additional bands, enhance cross-sensor transfer, and support arbitrary spectral configurations, \emph{AOM} introduces a \textit{Spectrum-independent Tokenizer} that performs channel-wise patch embedding augmented with channel index encoding, preserving per-band information while allowing nonlinear inter-band interactions in later layers. Second, to ensure robust feature capture across resolutions from sub-meter to hundred-meter scales, a \textit{Multi-scale Adaptive Patch Embedding} based on pseudo-inverse resize dynamically adjusts its receptive field to accommodate resolutions from sub-meter to hundred-meter, allowing \emph{AOM} to adapt its spatial granularity to the input resolution without sacrificing fine details or global context. Third, a self-supervised routine that masks and reconstructs individual channels teaches \emph{AOM} fine-grained spectral signatures and local spatial structure, yielding resilience to incomplete data. Finally, to maintain semantic consistency across diverse resolutions and imaging conditions, a multi-scale alignment constraint regularizes global representations. We validate \emph{AOM} on Geo-Bench and multiple mainstream datasets, including Sentinel-2, Landsat, and HLS. The results show that our model significantly outperforms existing models in downstream tasks, particularly in scenarios involving missing bands, cross-sensor data, and varying resolutions, showcasing its superior generalization. 

Overall, our contributions in this work are as follows: (1) We propose \emph{AOM}, a foundation model that adapts to arbitrary spectral channels, spatial resolutions, and sensor types, thereby overcoming the limitations of existing models. (2) We introduce spectrum-independent tokenizer to enhance cross-sensor generalization and a multi-scale patch embedding to enable resolution-adaptive feature extraction. (3) We develop channel-wise masked feature learning and reconstruction, alongside a global semantic alignment mechanism, to ensure multi-scale semantic consistency while improving the efficacy of self-supervised pretraining. (4) We validate \emph{AOM} on over 10 datasets and benchmarks, showing its superior performance and robustness in diverse scenarios.

\section{Related Work}

\subsubsection{Remote sensing foundation model.}
RSFMs are typically self-supervised networks pre-trained on large-scale satellite corpora and fine-tuned for diverse downstream tasks. Recent progress follows two main directions. One focuses on tailoring pretraining to RS-specific characteristics, such as multispectral inputs, small objects, and heterogeneous sensors. For instance, SatMAE~\cite{cong2022satmae} leverages masked autoencoding to preserve cross-band coherence, Skysense~\cite{guo2024skysense} fuses multi-source data with spatiotemporal cues via contrastive learning, SeaMo~\cite{li2025seamo} incorporates seasonal attributes, and AnySat~\cite{astruc2025anysat} adapts I-JEPA~\cite{assran2023self} to handle resolution diversity. The other line integrates vision-language models with geospatial priors: GeoChat~\cite{kuckreja2023geochat} uses instruction tuning for scene understanding, while GeoPixel~\cite{shabbir2025geopixel} introduces pixel-level decoding with adaptive partitioning. Despite their flexibility, these approaches primarily emphasize task-level customization, keeping core architectures static, potentially limiting generalization across the full spectrum of RS data diversity.
\begin{figure*}[ht]
      \centering	   
      \includegraphics[width=1.0\textwidth]{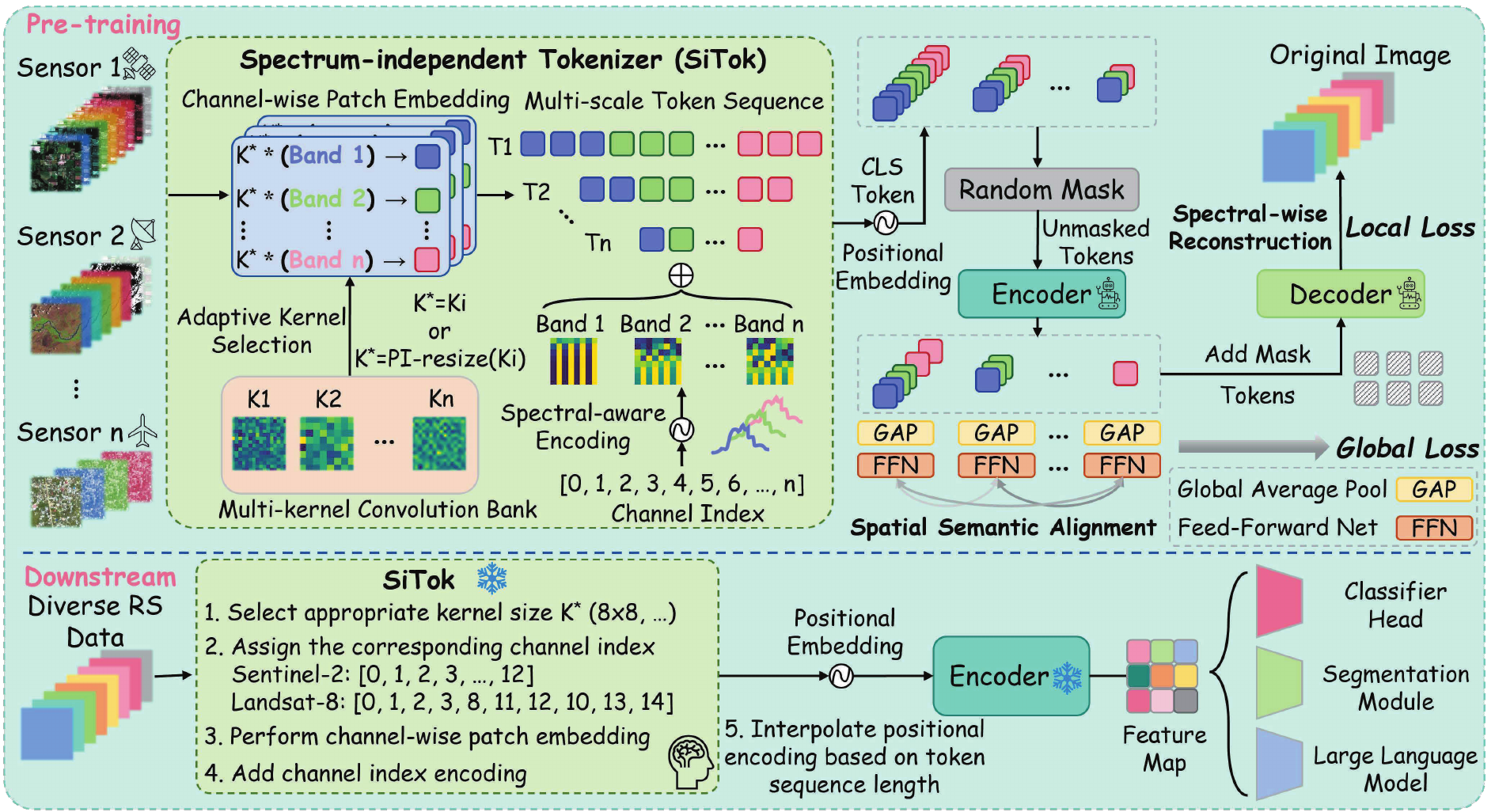}
      \caption{\textbf{An illustration of the proposed model.} AOM unifies spectral and spatial modeling through channel-wise patch embedding, adaptive multi-scale extraction, spectral-wise masking \& reconstruction, and multi-scale alignment, enabling flexible band configurations and robust performance across resolutions.}
\label{fig:workflow}
\end{figure*}
\subsubsection{Channel vision transformer.}
Vision Transformer (ViT)~\cite{dosovitskiy2020image} has become the core architecture for foundation models by leveraging self-attention for global image understanding. However, standard ViTs are limited to predefined settings such as natural RGB images and struggle with multi-spectral data, where spectral bands vary across sensors, as seen in biomedical and remote sensing domains. To address this, several channel-aware variants have been proposed. ChannelViT~\cite{baochannel2025} introduces per-channel tokens with hierarchical sampling; ChAda-ViT~\cite{bourriez2024chada} adds inter-channel attention for improved spatial interaction; DiChaViT~\cite{pham2024enhancing} employs a channel-diversity loss to retain band-specific features; and ChA-MAEViT~\cite{pham2025cha} enhances self-supervised learning through dynamic channel masking and memory tokens. While these methods improve multi-spectral handling, achieving broad adaptability across diverse imaging modalities remains unresolved.

\section{Methodology}
\subsection{Overview of the Proposed AOM}
To address the critical domain gap limitation in current RSFMs caused by spectral and spatial-scale discrepancies between pretraining and downstream tasks, we propose \emph{AOM}, a novel architecture with three key innovations. As described in Figure~\ref{fig:workflow}, we first design a spectrum-independent tokenizer that processes each spectral band individually through dedicated tokenization while incorporating channel index encoding for spectrum-aware representation. Second, to handle varying image resolutions and sizes, we develop a multi-scale patch embedding strategy based on pseudo-inverse resize transformation~\cite{beyer2023flexivit}, enabling adaptive patch representation across different spatial scales. Third, for effective large-scale pretraining, we enhance the Masked Autoencoder framework with two specialized components: (1) channel-wise reconstruction to capture spectral sequential properties, and (2) semantic consistency constraints across different scale embeddings.

\subsection{Spectrum-independent Tokenizer (SiTok)}

Conventional RSFMs typically process input images $\mathbf{I} \in \mathbb{R}^{C \times H \times W}$ with a fixed number of spectral channels $C$, limiting their flexibility for multi-modal applications. To enable robust processing of arbitrary spectral configurations, including handling missing or additional channels, we propose a novel \textit{Spectrum-independent Tokenizer} (\textbf{SiTok}) with two components.

\noindent
\textbf{\textit{Channel-wise patch embedding.}}
 For each spectral channel $\mathbf{S}_i \in \mathbb{R}^{1 \times H \times W}$, we apply a shared convolutional kernel $\mathbf{K} \in \mathbb{R}^{P \times P \times 1 \times D}$ to extract patch embeddings:
\(
\mathbf{I}_P = \{\mathbf{I}_1, \mathbf{I}_2, \dots, \mathbf{I}_C\}, \quad \text{where} \quad \mathbf{I}_i \in \mathbb{R}^{(N_H \times N_W) \times D}.
\)
Here, $P$ denotes the patch size, $D$ denotes the embedding dimension, and $N_H =  H/P $, $N_W =  W/P $ represent the spatial dimensions of the token grid. This design ensures consistent processing regardless of the input's channel count.

\noindent
\textbf{\textit{Spectral-aware encoding.}}
 We augment each channel's tokens with spectral-aware positional encodings:
\(
\mathrm{PE}_{\text{channel}} = \{\mathrm{PE}_1, \mathrm{PE}_2, \dots, \mathrm{PE}_C\},\quad \mathrm{PE}_i \in \mathbb{R}^{1 \times D},
\)
where each $\mathrm{PE}_i$ is generated via sinusoidal encoding of the channel index $i$. The encoding is then broadcast to $\mathbb{R}^{(N_H \times N_W) \times D}$ and added to the corresponding channel:
\(
\mathbf{I}_i + \mathrm{PE}_i \rightarrow \mathbf{I}_i\). This encoding preserves spectral ordering information while maintaining permutation invariance.

The final token sequence $\mathbf{T} \in \mathbb{R}^{(C \times N_H \times N_W) \times D}$ concatenates the tokens from all channels and forms a spectrally-aware representation suitable for transformer processing. Compared to fixed-channel tokenizers, our approach offers two key advantages: (1) inherent adaptability to varying channel configurations, (2) preservation of spectral characteristics through explicit encoding.

\subsection{Multi-scale Adaptive Patch Embedding (MAPE)}
\label{sec:mape}

Conventional patch embedding layers employ a single convolutional kernel
\(\mathbf{K}_{\mathrm{old}}\!\in\!\mathbb{R}^{P\times P\times C\times D}\)
with a fixed patch size~\(P\).
Since RS images span a broad range of spatial resolutions
and scene scales, a single \(P\) cannot capture both fine-grained
textures and coarse contextual structures.
Previous work~\cite{li2025fleximo}, therefore, resizes the kernel via the
pseudo-inverse resize (PI-resize) operator
\(\mathbf{K}_{\mathrm{new}}=\operatorname{PI\text{-}resize}%
(\mathbf{K}_{\mathrm{old}},r)\),
but large scale factors \(r\) introduce interpolation errors that degrade
performance.
To overcome this limitation, we introduce
\textit{Multi-scale Adaptive Patch Embedding} (\textbf{MAPE}), which extends
SiTok with the following components.

\noindent
\textbf{\textit{Multi-scale convolutional bank.}}
MAPE maintains a bank of \(n\) convolutions with different receptive
fields:
\(
\mathcal{K}\;=\;
\bigl\{\mathbf{K}_{1},\mathbf{K}_{2},\dots,\mathbf{K}_{n}\bigr\},\quad
\mathbf{K}_{i}\in\mathbb{R}^{P_{i}\times P_{i}\times C\times D},
\)
where \(P_{i}\) is the kernel size of the \(i\)-th branch.
Small \(P_{i}\) (\(P_{i}\!\!\rightarrow\!1\)) excels at capturing
high-frequency details, whereas large \(P_{i}\) (\(P_{i}\!\!\uparrow\))
gathers global context; the model thus adapts its computational footprint
and receptive field to the input.

\noindent
\textbf{\textit{Adaptive kernel selection.}}
For imagery that demands fine spatial detail, AOM employs a small
convolutional kernel; conversely, applications that emphasise global
context or higher throughput adopt a larger kernel.
Given a target patch size \(P_{t}\), the closest kernel in the bank is
\begin{equation}
i^{\star}\;=\;\arg\min_{i}\lvert P_{i}-P_{t}\rvert,
\end{equation}
and the effective weights are
\begin{equation}
\mathbf{K}^{\star}=
\begin{cases}
\mathbf{K}_{i^{\star}}, &
P_{i^{\star}} = P_{t},\\[6pt]
\operatorname{PI\text{-}resize}%
\bigl(\mathbf{K}_{i^{\star}},r\bigr), &
P_{i^{\star}}\neq P_{t},
\end{cases}
\quad
r=\frac{P_{t}}{P_{i^{\star}}}.
\end{equation}

\noindent
\textbf{\textit{PI-resize definition.}}
Let \(\boldsymbol{\omega}=\operatorname{vec}(\mathbf{K_i})\) vectorise the
\(P\times P\) spatial dimensions and
\(B^{r}\!\in\!\mathbb{R}^{(rP)^{2}\times P^{2}}\) denote the
bilinear-resize matrix.
PI-resize projects \(\boldsymbol{\omega}\) through the Moore–Penrose
pseudo-inverse of \(B^{r}\):
\begin{equation}
\operatorname{PI\text{-}resize}(\mathbf{K_i},r)\;=\;
\operatorname{reshape}\Bigl((B^{r\,\top})^{+}\,\boldsymbol{\omega}\Bigr),
\end{equation}
yielding the unique solution that minimises the expected response
distortion
\(
\mathbb{E}_{\mathbf{x}}\!
\bigl[\langle\mathbf{x},\mathbf{K_i}\rangle-
      \langle B^{r}\mathbf{x},\mathbf{K_i}'\rangle\bigr]^{2}
\).
Because PI-resize is applied only to the nearest kernel, the approximation error is bounded, ensuring stable feature extraction for
arbitrary \(P_{t}\).

\noindent
\textbf{\textit{Integration with SiTok.}}
\textbf{MAPE} is a plug-and-play module that can augment any tokenizer.
To insert it into \textbf{SiTok} we instantiate the multi-scale convolutional bank for a \mbox{single-channel} input,
\(
\mathcal{K}
   =\{\mathbf{K}_{1},\mathbf{K}_{2},\dots,\mathbf{K}_{n}\},
   \mathbf{K}_{i}\in\mathbb{R}^{P_{i}\times P_{i}\times 1\times D}.
\)
Adaptive kernel selection provides the
resolution-matched weights \(\mathbf{K}^{\star}\), which are convolved
with the input image \(\mathbf{I}\) to yield
channel-wise patch tokens; these tokens are subsequently enriched
by the \emph{Spectral-aware Encoding} inherited from \textbf{SiTok}.

\subsection{Semantic Alignment Pretraining Task}
The data-ingest module of \emph{AOM} couples \textbf{SiTok} with our
\textbf{MAPE}.
To exploit large-scale RS corpora without manual
labels, We design a dual-objective self-supervised approach combining Masked Autoencoding and Contrastive Learning to capture both local spatial-spectral patterns and global semantic alignment.

For an input image $\mathbf{I} \in \mathbb{R}^{C \times H \times W}$, the \textbf{SiTok} and \textbf{MAPE} modules generate $n$ parallel token sequences at different scales:
\(
\{\mathbf{T}_1, \mathbf{T}_2, \dots, \mathbf{T}_n\}, \quad \text{where} \quad \mathbf{T}_i \in \mathbb{R}^{(N_i \times D)}, 
\)
where $N_i = C \times (H/P_i) \times (W/P_i)$ and $P_i$ denotes the patch size at scale $i$. We then randomly mask portions of each token sequence with masking ratio $m \in (0,1)$, preserving only the unmasked tokens $\mathbf{T}_i^{\text{vis}} \in \mathbb{R}^{(N_i \times (1-m)) \times D)}$ for encoder processing:
\begin{equation}
\begin{aligned}
\mathbf{E}_i = \text{Encoder}(\mathbf{T}_i^{\text{vis}}), \quad \\
\text{where} \quad \mathbf{E}_i \in \mathbb{R}^{(N_i \times (1-m)) \times D}.
\end{aligned}
\end{equation}
The encoded features undergo parallel optimization through two complementary learning objectives:

\noindent
\textbf{\textit{Masked spectral–spatial reconstruction.}}
For every scale~$i$ we instantiate an individual decoder $\operatorname{Decoder}_i$.
We first augment the encoder output $\mathbf{E}_i$ with learnable mask tokens
$\mathbf{M} \in \mathbb{R}^{(N_i \times m) \times D}$ and then feed the concatenated
sequence into the corresponding decoder:
\(
    \hat{\mathbf{T}}_i = \operatorname{Decoder}_i\bigl(\operatorname{Concat}(\mathbf{E}_i, \mathbf{M})\bigr).
\)
The reconstruction loss is the pixel-wise mean-squared error (MSE) between the
predicted and ground-truth patches across all spectral channels:
\begin{equation}
    \mathcal{L}_{\text{Recon}} =
    \frac{1}{n}\sum_{i=1}^n
    \bigl\|\hat{\mathbf{T}}_i - \mathbf{T}_i^{\mathrm{masked}}\bigr\|_2^2. 
\end{equation}

\noindent
\textbf{\textit{Multi-scale semantic alignment.}} To ensure that tokens derived from different patch scales encode
consistent semantics, we introduce a
multi-scale alignment branch. We apply global average pooling (GAP) followed by non-linear projection heads $g_i(\cdot)$:
\begin{equation}
    \mathbf{H}_i = g_i(\text{GAP}(\mathbf{E}_i)), \quad \mathbf{H}_i \in \mathbb{R}^d 
\end{equation}
    The alignment loss encourages feature similarity across scales using InfoNCE:
\begin{equation}
    \mathcal{L}_{\text{Align}} = -\sum_{i=1}^n \sum_{\substack{j=1 \\ j \neq i}}^n \log \frac{\exp(s(\mathbf{H}_i, \mathbf{H}_j)/\tau)}{\sum_{k=1}^n \exp(s(\mathbf{H}_i, \mathbf{H}_k)/\tau)}
\end{equation}
where $s(\cdot,\cdot)$ denotes cosine similarity and $\tau$ is a temperature parameter.

\noindent
\textbf{\textit{Composite objective function.}} 
The total training objective combines both losses with balanced weighting:
\begin{equation}
\mathcal{L}_{\text{Total}} = \lambda_1 \mathcal{L}_{\text{Recon}} + \lambda_2 \mathcal{L}_{\text{Align}}
\end{equation}
where $\lambda_1$ and $\lambda_2$ control the relative importance of reconstruction fidelity versus cross-scale semantic alignment.
This dual supervision compels \emph{AOM} to model local
spectral-spatial correlations through channel-wise reconstruction while
simultaneously aligning global semantics across multiple patches
scales, providing strong generalization for downstream RS tasks.

\begin{table*}[!t]
\centering
\setlength{\tabcolsep}{3pt}
\renewcommand{\arraystretch}{1}
\begin{tabular}{lccccccc}
\toprule
\multirow{3}{*}{\textbf{Model}} & m-pv4ger-seg & m-nz-cattle & m-neonTree & m-cashew-plant. & m-SA-crop-type & m-chesapeake&\multirow{3}{*}{\textbf{Overall}} \\[-0.2em]
 & \makecell{(RGB)\\$320\times320$}
 & \makecell{(RGB)\\$500\times500$}
 & \makecell{(RGB)\\$400\times400$}
 & \makecell{(Sentinel-2)\\$256\times256$}
 & \makecell{(Sentinel-2)\\$256\times256$}
 & \makecell{(RGBN)\\$256\times256$} & \\ 
\midrule
Random init.      & 76.6 & 67.5 & 46.9 & 37.1 & 26.7 & 38.9 & 48.95 \\
MAE (single)      & 85.4 & 70.1 & 53.0 & 45.7 & 27.5 & 42.1 & 53.96 \\
Scale-MAE         & 78.4 & 72.7 & 51.0 & —   & —   & 48.9 & —     \\
GFM               & 84.8 & 71.8 & 51.1 & —   & —   & 54.1 & —     \\
SenPaMAE          & 81.3 & 75.3 & 50.5 & 48.7 & 28.1 & 51.7 & 55.93 \\
CROMA             & —   & —   & —   & \underline{55.6} & \textbf{31.4} & —   & —     \\
AnySat            & \underline{90.2} & 76.9 & 51.8 & 52.3 & 29.5 & \underline{55.3} & 59.33 \\
DOFA              & 89.1 & \underline{77.8} & \underline{53.2} & 53.8 & 29.0 & 53.8 & \underline{59.45} \\
\midrule
\textbf{AOM (ours)} 
  & \textbf{91.5}\,{\scriptsize(+1.3)} 
  & \textbf{80.2}\,{\scriptsize(+2.4)} 
  & \textbf{53.7}\,{\scriptsize(+0.5)} 
  & \textbf{68.3}\,{\scriptsize(+12.7)} 
  & \underline{31.0} 
  & \textbf{59.2}\,{\scriptsize(+3.9)} 
  & \textbf{63.98}\,{\scriptsize(+4.53)} \\
\bottomrule
\end{tabular}
\caption{Partial fine-tuning results (mIoU) on the six Geo-Bench semantic-segmentation tasks.  
All backbones are frozen, and a UPerNet head is trained for 20 epochs.  
\textbf{Bold} numbers denote the best performance, \underline{underlined} numbers mark the second-best, and the values in parentheses indicate the absolute improvement of our method over the respective second-best model.}
\label{tab:Geo-Bench}
\end{table*}
\subsection{Apply AOM on Diverse RS Data}
After pretraining, \emph{AOM} is deployed to diverse optical RS datasets (Fig.~\ref{fig:workflow}). Given the spatial resolution and image size, we first select a suitable patch-embedding kernel to balance detail and computation. We then assign channel indices according to the sensor’s spectral configuration (e.g., Sentinel-2’s 13 bands are indexed 0–12, and RGB or Landsat data select their corresponding indices). Since \emph{AOM} models channels independently, missing or newly added bands do not affect its ability to exploit previously learned ones. Patch embedding is applied with the chosen kernel, channel-index encoding is added, and the positional encoding is interpolated to match the resulting token sequence. These tokens are finally fed into the pretrained encoder for feature extraction in downstream tasks such as classification, segmentation, or as visual input to a large language model.
\section{Experiments}
Our experimental section begins with the pretraining configuration detailing the multi-source datasets and optimization strategy. We then comprehensively assess \emph{AOM} on the Geo-Bench benchmark and diverse cross-sensor datasets, demonstrating superior performance across all metrics. Rigorous ablation studies further validate the model's robustness to varying image sizes, spatial resolutions, and spectral configurations, consistently outperforming existing RSFMs across all test scenarios.
\subsection{Pretraining Configurations}
\noindent
\textbf{\textit{Pretraining dataset.}}
Our pretraining dataset comprises multi-source remote sensing imagery carefully selected to enable the model to learn cross-sensor characteristics during pretraining. We employ a diverse collection of multi-sensor remote sensing datasets for pretraining, including: (1) Sentinel-2 imagery from SSL4EO-S12 (1.004 million samples at 10-60m resolution)~\cite{wang2023ssl4eo}, (2) Landsat-8 data from Activefire (146k samples at 30-100m resolution)~\cite{de2021active}, and (3) high-resolution RGB collections from GeoPile~\cite{mendieta2023towards}, fMoW~\cite{christie2018functional}, and OpenEarthMap~\cite{xia2023openearthmap} (108k samples at 0.1-30m resolution). This combined dataset spans resolutions from 0.1m to 100m, totaling approximately 1.56 million samples across optical, multispectral, and high-resolution domains to ensure a comprehensive representation of remote sensing scenarios.
\begin{table}[!t]
\centering
\setlength{\tabcolsep}{4pt}
\renewcommand{\arraystretch}{1.0}
\begin{tabular}{lcc}
\toprule
{\textbf{Model}}& \makecell{SPARCS\\(Landsat-8)} & \makecell{HLS Burn Scars\\(HLS)}\\
\midrule
SatMAE (NeurIPS’22)          & 49.9 & 81.1 \\
CROMA (NeurIPS’23)           & 52.3 & \underline{82.4} \\
SpectralGPT (TPAMI’24)    & \underline{57.6} & 80.5 \\
DOFA (ArXiv’24)           & 55.4 & 80.6 \\
SeaMo (INF’25)          & 51.7 & 81.8 \\
\midrule
\textbf{AOM (ours)}     & \textbf{68.5}\,{\scriptsize(+10.9)} & \textbf{85.4}\,{\scriptsize(+3.0)} \\
\bottomrule
\end{tabular}
\caption{Partial fine-tuning results (mIoU) on two cross-sensor segmentation tasks.
\textbf{Bold} marks the best, \underline{underline} denotes the second-best, and parentheses report AOM’s absolute gain over the latter.}
\label{tab:cross_sensor_finetune}
\end{table}
\begin{table}[ht]
\centering
\setlength{\tabcolsep}{4pt}
\renewcommand{\arraystretch}{1.0}
\begin{tabular}{lcc}
\toprule
\textbf{Model} & \makecell{UCM\\(RGB)} & \makecell{BigEarthNet\\(Sentinel-2)} \\
\midrule
SatMAE (NeurIPS’22)     & 85.17 & 79.36 \\
CROMA (NeurIPS’23)      & —    & \underline{83.41} \\
SpectralGPT (TPAMI’24)  & 82.42 & 81.05 \\
DOFA (ArXiv’24)         & \underline{90.09} & 82.45 \\
AnySat (CVPR’25)        & 88.92 & 82.79 \\
\midrule
\textbf{AOM (ours)}   & \textbf{93.57}\,{\scriptsize(+3.48)} & \textbf{85.02}\,{\scriptsize(+1.61)} \\
\bottomrule
\end{tabular}
\caption{Linear probing (LP) accuracy (\%) on two classification datasets: overall accuracy on UCM and mAP on BigEarthNet. \textbf{Bold} marks the best score, \underline{underline} the second-best; numbers in parentheses give AOM’s absolute gain over the latter.}
\label{tab:linear_results}
\end{table}
\begin{figure}[!t]
      \centering	   
      \includegraphics[width=0.48\textwidth]{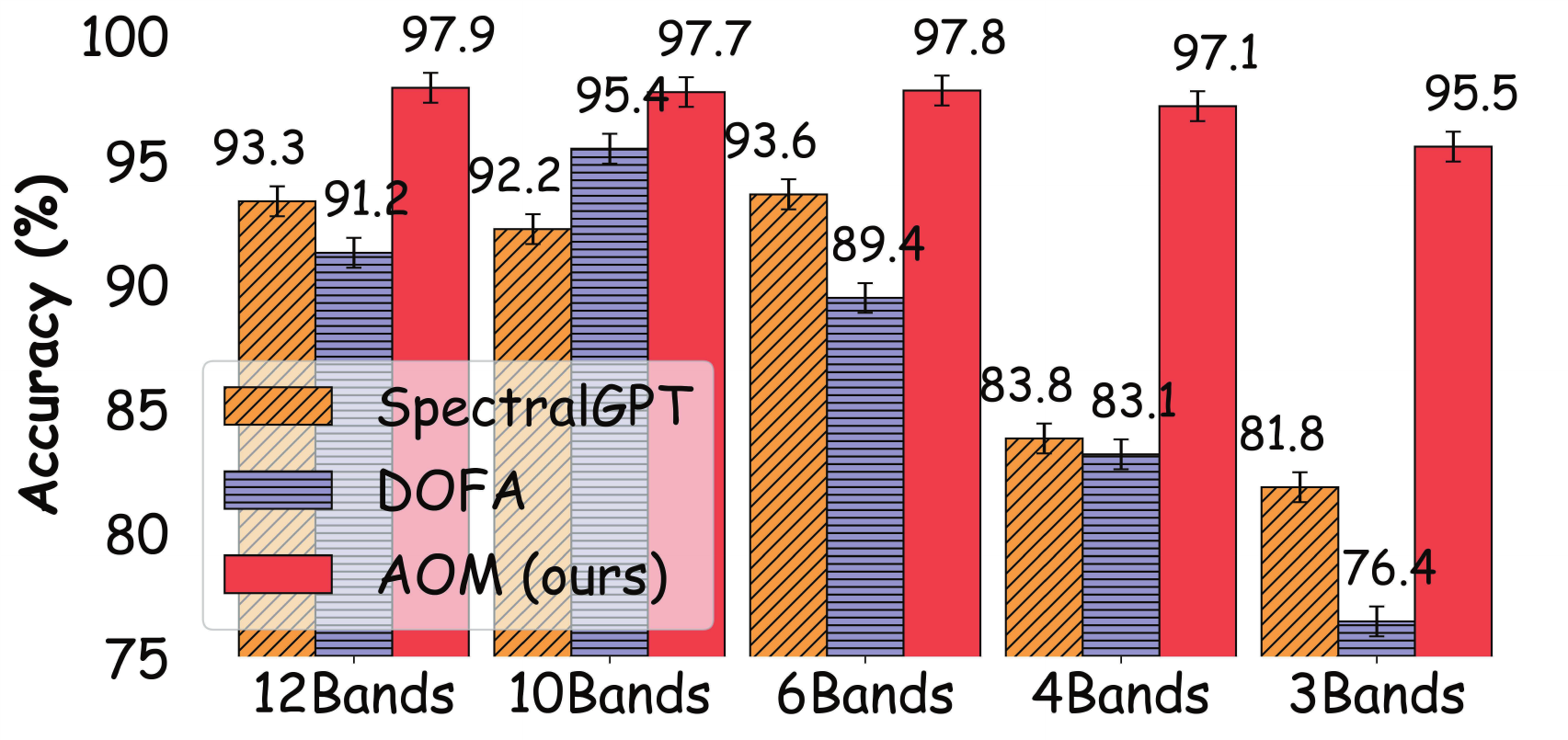}
      \caption{\textbf{Linear probing accuracy (\%) on EuroSAT across different band combinations.} EuroSAT images are captured by Sentinel-2 and contain 13 spectral bands (indexed 0–12). The y-axis represents classification accuracy, while the x-axis indicates the number of spectral bands used during training.}
\label{fig:channel}
\end{figure}

\noindent
\textbf{\textit{Pretraining details.}} 
During the pertaining, we run 220 epochs on the pretraining corpus with a batch size of~1024 and a base learning rate of~$1\times10^{-4}$.  
The multi-scale convolutional bank is initialized with kernel sizes $\{16,32,64\}$, while training sequentially cycles through patch sizes $\{16,24,32,48,64\}$, hence, five independent decoders are employed for reconstruction.  
Our \emph{adaptive kernel selection} mechanism dynamically resizes each convolutional kernel so that its receptive field is always aligned with the currently sampled patch size. The encoder of \emph{AOM} is based on the ViT-Base architecture, with 4 decoder layers applied. The image masking ratio is set to 75\%. For the loss function, since the multi-scale features extracted by \emph{AOM} are inherently positive pairs, we set the temperature parameter of the InfoNCE loss to 0.5 to strengthen similarity learning. Furthermore, because the overall magnitude of InfoNCE loss is lower than that of MSE loss, we assign weights of 0.8 and 0.2 to the InfoNCE and MSE losses, respectively, to balance their contributions during training. For data augmentation, we employ only simple augmentations, including random horizontal flipping and random cropping. Notably, the cropped images are resized back to the original sizes of each dataset. This means that the input image size during pretraining is not fixed, but rather varies according to the native resolution of each dataset image.
\begin{figure}[!t]
      \centering	   
      \includegraphics[width=0.45\textwidth]{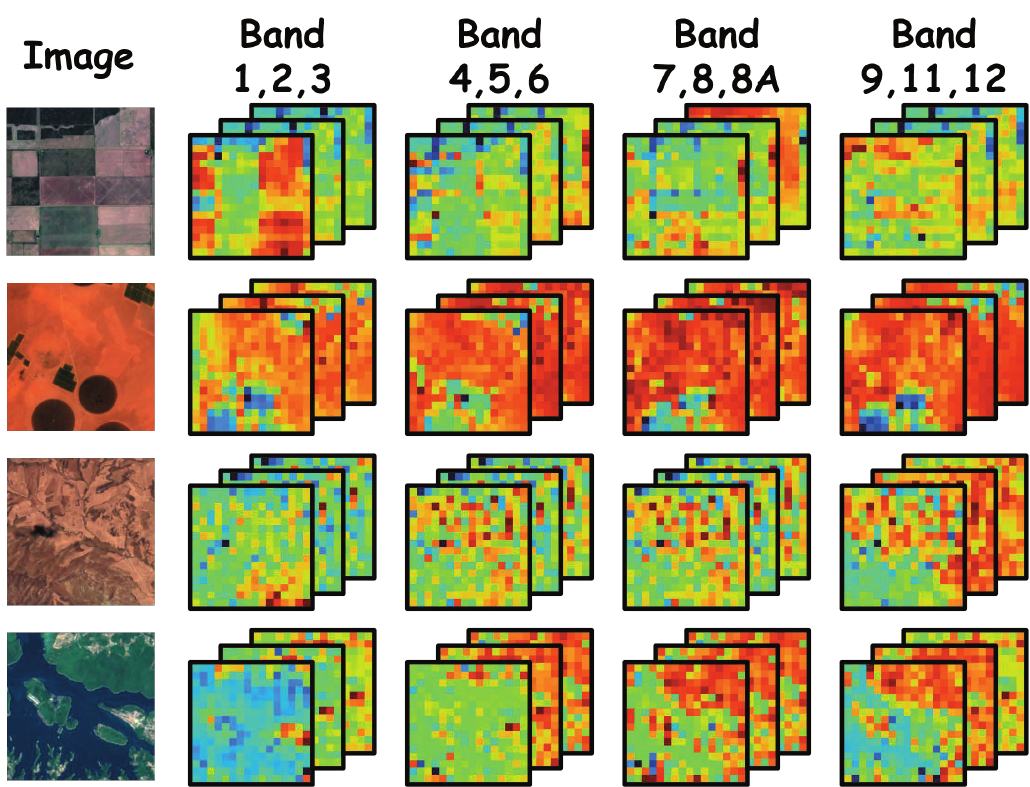}
      \caption{\textbf{Visualization of feature maps from different Sentinel-2 spectral bands through SiTok.} The results demonstrate the model's ability to extract features independently from each band.}
\label{fig:heatmap}
\end{figure}
\subsection{Downstream Evaluation}
\noindent
\textbf{\textit{Geo-Bench evaluation.}} 
We benchmark our model on the six semantic-segmentation datasets of \textbf{Geo-Bench}~\cite{lacoste2023geo}, a suite that deliberately covers multiple sensors.  Following the official protocol, we freeze the backbone for every method and train a UPerNet segmentation head~\cite{xiao2018unified} for 20 epochs. As shown in Table~\ref{tab:Geo-Bench}, our approach establishes new state-of-the-art (SOTA) results on five out of six datasets and delivers sizable gains over the second-best model; for instance, it improves the cashew-plantation score by \textbf{+12.7~mIoU}.  These findings underscore the proposed model’s robustness to both resolution changes and cross-sensor shifts.  Complete training details and additional metrics are reported in the supplementary material.

\noindent
\textbf{\textit{Diverse cross-sensor datasets.}} 
While Geo-Bench primarily covers segmentation tasks for RGB and Sentinel-2 imagery, many RSFMs are evaluated across various satellite datasets and tasks. To evaluate our model's generalization capability across different optical sensors, we conduct experiments on two representative datasets:
(1) \textbf{SPARCS}: Landsat-8 cloud and cloud shadow detection~\cite{hughes2019high}. (2) \textbf{HLS Burn Scars}: Harmonized Landsat-Sentinel (HLS) burn scar identification~\cite{HLS_Foundation_2023}. Following standard practice for fair comparison with existing RSFMs, we adopt a partial fine-tuning strategy where only the segmentation head is fine-tuned. This protocol effectively evaluates the model's transfer learning capability.
As shown in Table~\ref{tab:cross_sensor_finetune}, our model achieves significant improvements over existing methods on both datasets, demonstrating superior cross-sensor generalization. 

To assess our model's representation learning capability, we evaluate through linear probing (LP) on two standard RS classification benchmarks: (1) \textbf{UCM}: RGB land use classification~\cite{yang2010bag}. (2) \textbf{BigEarthNet}: Multi-label Sentinel-2 scene classification~\cite{sumbul2021bigearthnet}. We freeze the backbone weights and only train a linear classifier head. This LP scheme effectively measures the quality of learned representations without fine-tuning. As shown in Table~\ref{tab:linear_results}, our model achieves SOTA performance on both datasets, demonstrating superior representation learning capability across different sensor modalities. More data information, experimental results, implementation details, and visualizations are provided in the supplementary material.
\begin{figure}[!t]
      \centering	   
      \includegraphics[width=0.48\textwidth]{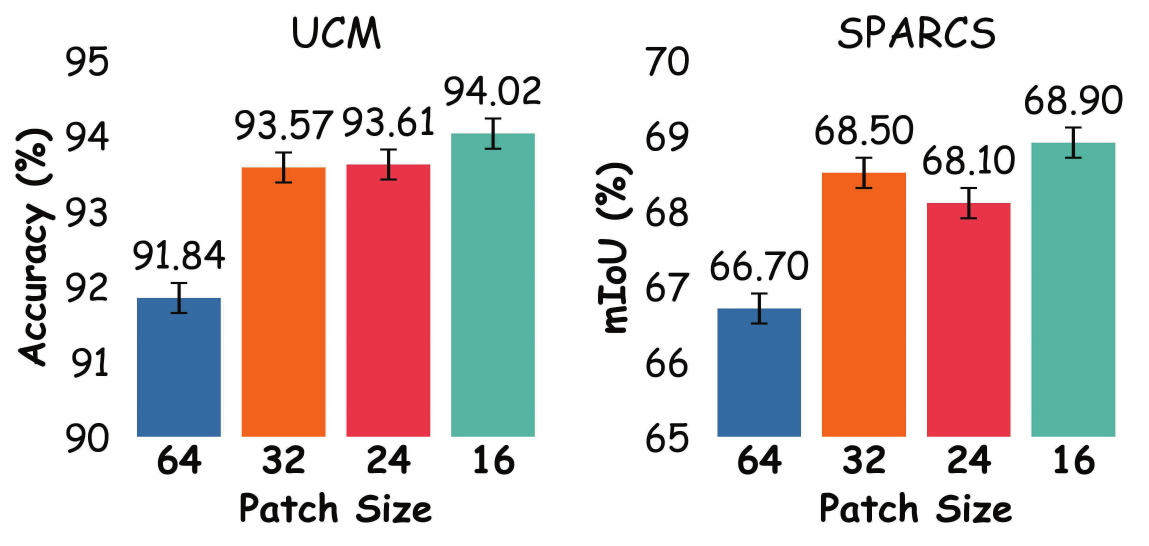}
      \caption{\textbf{Results on two datasets across different patch size configurations.} AOM maintains stable accuracy and mIoU across patch granularities, demonstrating robust performance in both classification and segmentation.}
\label{fig:patch size}
\end{figure}
\subsection{Ablation Studies}
\noindent
\textbf{\textit{Robustness to arbitrary band combinations.}} 
As mentioned in the introduction, existing RSFMs suffer a significant drop in feature extraction capability when the spectral bands in downstream tasks differ from those used during pretraining. To evaluate this issue, we freeze the backbone of \emph{AOM} and fine-tune only the classification head under various band combinations of Sentinel-2. \emph{AOM} demonstrates superior robustness across arbitrary band combinations in EuroSAT~\cite{helber2019eurosat} linear probing, consistently outperforming existing methods under all tested configurations. As shown in Figure~\ref{fig:channel}, it achieves 2.28-19.09\% higher accuracy than competitors, maintaining 95.50\% accuracy even with only 3 available bands, while others degrade significantly. The model shows particular strength in challenging scenarios, proving its unique capability to handle flexible spectral inputs without performance compromise. Figure~\ref{fig:heatmap} shows the visualized feature maps of Sentinel-2 images processed by \textbf{SiTok}, illustrating the model’s ability to extract distinct features from each channel.

\begin{figure}[!t]
      \centering	   
      \includegraphics[width=0.45\textwidth]{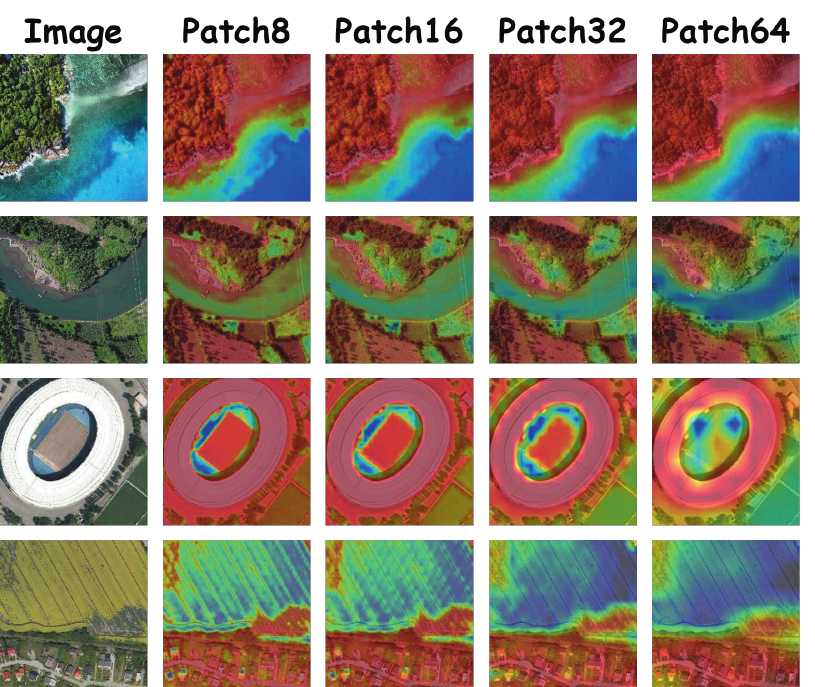}
      \caption{\textbf{Visualizing key regions of interest from AOM's final output feature map.} The results show that AOM preserves global semantics across different patch sizes.}
\label{fig:spatialmap}
\end{figure}

\noindent
\textbf{\textit{Ablation on patch sizes.}}
As shown in Figure~\ref{fig:patch size}, across the partial fine-tuning patch-size ablation on UCM and SPARCS, \emph{AOM}’s accuracy and mIoU remain remarkably stable as the patch granularity shifts from coarse to fine, underscoring the model’s ability to preserve performance and hence robustness over a wide range of spatial resolutions for classification and segmentation tasks. Figure~\ref{fig:spatialmap} visualizes the patch embeddings produced by \emph{AOM} at multiple scales.

\noindent
\textbf{\textit{Ablation on local vs.\ global semantic objectives.}} 
To disentangle the effect of the loss components, we compare (i) a local per-spectral reconstruction objective that minimises the patch-wise MSE, and (ii) the same MSE term augmented with an InfoNCE loss that enforces multi-scale, global semantic consistency across patches.  Table~\ref{tab:loss_ablation} shows that adding the InfoNCE term consistently boosts accuracy on all three datasets.  
This confirms that local spectral fidelity alone is insufficient: incorporating global, cross-scale cues produces stronger and more transferable representations.
\begin{table}[ht]
\centering
\setlength{\tabcolsep}{1pt}      
\renewcommand{\arraystretch}{1.15} 
\footnotesize                    
\begin{tabular}{lcccc}
\toprule
\multirow{2}{*}{\textbf{Loss}} 
&  \multirow{2}{*}{\textbf{Effects}}& \multicolumn{2}{c}{\textbf{Acc.\,(\%)$\uparrow$}} 
  & \textbf{mIoU\,(\%)$\uparrow$} \\
\cmidrule(lr){3-4}\cmidrule(lr){4-5}
& & EuroSAT & UCM & SPARCS \\
\midrule
MSE          &Local& 96.45 & 91.28 & 66.7 \\
MSE\&InfoNCE  &Local\&Global& \textbf{97.87} 
                         & \textbf{93.57} 
                         & \textbf{68.5} \\
\bottomrule
\end{tabular}
\caption{Effect of loss design on model performance.}
\label{tab:loss_ablation}
\end{table}

\noindent
\textbf{\textit{Spectral encoding ablation.}} 
To incorporate prior information into the channel token sequence after patch embedding, several strategies can be considered. One approach involves embedding the central wavelength of each spectral band, while another encodes the channel index in sequential order (i.e., 0, 1, 2, 3, ...). Though prior works have typically adopted wavelength embeddings, our experiments reveal minimal performance differences between the two methods, as described in Figure~\ref{tab:encoding_ablation}. Given its simplicity and ease of use, particularly in downstream applications where exact wavelength metadata may be unavailable, we adopt channel index encoding as a more practical alternative.
\begin{table}[ht]
\centering
\setlength{\tabcolsep}{3pt}      
\renewcommand{\arraystretch}{1.15} 
\footnotesize                    
\begin{tabular}{lccc}
\toprule
\multirow{2}{*}{\textbf{Encoding settings}} 
  & \multicolumn{2}{c}{\textbf{Acc.\,(\%)$\uparrow$}} 
  & \textbf{mIoU\,(\%)$\uparrow$} \\
\cmidrule(lr){2-3}\cmidrule(lr){4-4}
& EuroSAT & UCM & SPARCS \\
\midrule
Wavelength     & 97.72 & 92.81 & 67.7 \\
Channel Index  & \textbf{97.87} 
                         & \textbf{93.57} 
                         & \textbf{68.5} \\
\bottomrule
\end{tabular}
\caption{Effect of spectral encoding on model performance.}
\label{tab:encoding_ablation}
\end{table}

\section{Conclusion}
In this paper, we propose Any-Optical-Model (AOM), a universal foundation model for remote sensing designed to handle arbitrary spectral bands, resolutions, and image sizes. The key innovations of AOM are: (1) a spectrum-independent tokenizer that ensures adaptability to varying band configurations while preserving spectral information; (2) a multi-scale adaptive patch embedding mechanism that dynamically adjusts to diverse spatial scales; and (3) a self-supervised pretraining strategy with channel-wise masking and multi-scale semantic alignment to enhance robustness and consistency. We conducted extensive experiments on over 10 datasets demonstrating that AOM outperforms existing RSFMs in downstream tasks, particularly under challenging conditions such as missing bands, cross-sensor data, and varying resolutions. These results establish AOM as a significant advancement toward truly universal RSFMs.

\textbf{Limitations and future work.} Although existing studies on AOM have demonstrated its effectiveness across multiple sensors and tasks, there remain two key challenges regarding its generalization and universality. First, its robustness to out-of-distribution data such as hyperspectral imagery and SAR still requires rigorous evaluation. Second, its performance on a broader range of tasks, including object detection and temporal prediction, remains underexplored. To this end, future work will extend AOM to more diverse sensors and tasks, investigate more effective channel-embedding strategies, and further examine its capability to extract features across a wider range of spatial resolutions.

\section{Acknowledgements}
This work was supported by the National Key Research and Development Program of China under Grant 2022YFB3903401, the National Natural Science Foundation of China under Grant 42271350, and by the International Partnership Program of the Chinese Academy of Sciences under Grant No.313GJHZ2023066FN.

\bibliography{aaai2026}
\end{document}